\documentclass[graybox]{svmult}

\usepackage{mathptmx}       
\usepackage{helvet}         
\usepackage{courier}        
\usepackage{type1cm}        
\usepackage{makeidx}         
\usepackage{graphicx}        
\usepackage{multicol}        
\usepackage[para]{footmisc} 
\usepackage{xcolor}
\usepackage{array}
\usepackage{multirow} 
\usepackage{caption}
\usepackage{listings} 
\usepackage[toc,page]{appendix}

\usepackage{titlesec}
\titlespacing\section{0pt}{12pt plus 4pt minus 2pt}{0pt plus 2pt minus 2pt}
\titlespacing\subsection{0pt}{8pt plus 4pt minus 2pt}{0pt plus 2pt minus 2pt}

\makeindex             

\newcommand{\Xingkun}[1]{{\color{violet} XL: #1}}
\newcommand{\XingkunDel}[1]{}
\newcommand{\VR}[1]{{\color{magenta} VR: #1}}

\newcommand{\VRdel}[1]{}

\colorlet{punct}{red!60!black}
\definecolor{background}{HTML}{EEEEEE}
\definecolor{delim}{RGB}{20,105,176}
\colorlet{numb}{magenta!60!black}
\lstdefinelanguage{json}{
    basicstyle=\normalfont\ttfamily,
    numbers=left,
    numberstyle=\scriptsize,
    stepnumber=1,
    numbersep=8pt,
    showstringspaces=false,
    breaklines=true,
    frame=lines,
    literate=
     *{0}{{{\color{numb}0}}}{1}
      {1}{{{\color{numb}1}}}{1}
      {2}{{{\color{numb}2}}}{1}
      {3}{{{\color{numb}3}}}{1}
      {4}{{{\color{numb}4}}}{1}
      {5}{{{\color{numb}5}}}{1}
      {6}{{{\color{numb}6}}}{1}
      {7}{{{\color{numb}7}}}{1}
      {8}{{{\color{numb}8}}}{1}
      {9}{{{\color{numb}9}}}{1}
      {:}{{{\color{punct}{:}}}}{1}
      {,}{{{\color{punct}{,}}}}{1}
      {\{}{{{\color{delim}{\{}}}}{1}
      {\}}{{{\color{delim}{\}}}}}{1}
      {[}{{{\color{delim}{[}}}}{1}
      {]}{{{\color{delim}{]}}}}{1},
}

\begin{document}

\title*{Benchmarking Natural Language Understanding Services for building Conversational Agents}
\titlerunning{Benchmarking NLU Services for building Conversational Agents}

\author{Xingkun Liu, Arash Eshghi, Pawel Swietojanski and Verena Rieser}


\institute{Xingkun Liu, Arash Eshghi and Verena Rieser \at Heriot-Watt University, Edinburgh, EH14 4AS, \email{\url{[x.liu], [a.eshghi], [v.t.rieser]@hw.ac.uk}}
\and Pawel Swietojanski \at The University of New South Wales, Sydney, Australia,  \email{\url{p.swietojanski@unsw.edu.au}} (work done when Pawel was with Emotech North LTD)\\}

%
%

\maketitle

\abstract{We have recently seen the emergence of several publicly available Natural Language Understanding (NLU) toolkits, which map user utterances to structured, but more abstract, Dialogue Act (DA) or Intent specifications, while making this process accessible to the lay developer. 
In this paper, we present the first  wide coverage evaluation and comparison of 
 some of the most popular NLU services, on a large, multi-domain (21 domains) dataset of 25K user utterances
that we have collected and annotated with Intent and Entity Type specifications and which will be released as part of this submission.\footnote{\url{https://github.com/xliuhw/NLU-Evaluation-Data}} 
The results show that on Intent classification Watson significantly outperforms the other platforms, namely, Dialogflow, LUIS and Rasa; though these also perform well. Interestingly, on Entity Type recognition, Watson performs significantly worse due to its low Precision\footnote{At the time of producing the camera-ready version of this paper, we noticed the seemingly recent addition of a `Contextual Entity' annotation tool to Watson, much like e.g. in Rasa. We'd threfore like to stress that this paper does \emph{not} include an evaluation of this feature in Watson NLU}. Again, Dialogflow, LUIS and Rasa perform well on this task. 
}

\section{Introduction}
\label{sec:1}
Spoken Dialogue Systems (\VRdel{henceforth}SDS), or Conversational Agents are ever more common in home and work environments, and the market is only expected to grow. This has prompted industry and academia to create platforms for fast development of SDS, with interfaces that are designed to make this process easier and more accessible to those without expert knowledge of this multi-disciplinary research area. 

One of the key SDS components for which there are now several such platforms available is the Natural Language Understanding (NLU) component, which maps individual utterances to structured, abstract representations, often called Dialogue Acts (DAs) or Intents together with their respective arguments that are usually Named Entities within the utterance. Together, the representation is taken to specify the semantic content of the utterance as a whole in a particular dialogue domain.

In the absence of reliable, third-party -- and thus unbiased -- evaluations of NLU toolkits, it is difficult for users (which are often conversational AI companies) to choose between these platforms. In this paper, our goal is to provide just such an evaluation: we present the first systematic, wide-coverage evaluation of some of the most commonly used\footnote{according to anecdotal evidence from academic and start-up communities} NLU services, namely: Rasa\footnote{\url{https://rasa.com/}}, Watson\footnote{\url{https://www.ibm.com/watson/ai-assistant/}}, LUIS\footnote{\url{https://www.luis.ai/home}} and Dialogflow\footnote{\url{https://dialogflow.com/}}. 
The evaluation uses a new dataset of 25k user utterances which we annotated with Intent and Named Entity specifications. 
The dataset, as well as our evaluation toolkit will be released for public use.

\section{Related Work}

To our knowledge, 
this is the first wide coverage comparative evaluation  of  NLU services - those that exist tend to lack breadth in Intent types, Entity types, and the domains studied. For example, recent blog posts  \cite{Wisniewski.etal17,Coucke.etal17}, summarise benchmarking results for 4 domains, with only 4 to 7 intents for each.
The closest published work to the results presented here is by
\cite{Braun.etal17}, who evaluate 6 NLU services in terms of their accuracy (as measured by precision, recall and F-score, as we do here) on 3 domains with 2, 4, and 7 intents and 5, 3, and 3 entities respectively.
In contrast, we consider the 4 currently most commonly used NLU services on a large, new data set, which contains 21 domains of different complexities, covering 64 Intents and 54 Entity types in total.  
In addition, \cite{Massimo-Canonico.etal18} describe an analysis of NLU engines in terms of their usability, language coverage, price etc., which is complimentary to the work presented here.


\section{Natural Language Understanding Services}
\label{sec:3}
There are several options for building the NLU component for conversational systems. NLU typically performs the following tasks: (1) Classifying the user Intent or Dialogue Act type; and (2) Recognition\XingkunDel{\& Classification} of Named Entities (henceforth NER) in an utterance\footnote{Note that, one could develop one's own system using existing libraries, e.g. sk\_learn libraries \url{http://scikit-learn.org/stable/}, spaCy \url{https://spacy.io/}, but a quicker and more accessible way is to use an existing service platform.}. 
%
There are currently a number of service platforms that perform (1) and (2): commercial ones, such as Google's Dialogflow (formerly Api.ai), 
 Microsoft's LUIS, IBM's Watson Assistant (henceforth Watson), Facebook's Wit.ai, Amazon Lex, Recast.ai, Botfuel.io; and open source ones, such as Snips.ai\footnote{was not yet open source when we were doing the benchmarking, and was later on also introduced in \url{https://arxiv.org/abs/1805.10190}} and Rasa.
As mentioned above, we focus on four of these: Rasa, IBM's Watson, Microsoft's LUIS and Google's Dialogflow. 
In the following, we briefly summarise and discuss their various features. Table \ref{nlu-services-inuputs-outputs} provides a summary of the input/output formats for each of the platforms.

(1) All four platforms support Intent classification and NER; (2) None of them support Multiple Intents where a single utterance might express more than one Intent, i.e. is performing more than one action. This is potentially a significant limitation because such utterances are generally very common in spoken dialogue; (3) Particular Entities and Entity types tend to be \emph{dependent} on particular Intent types, e.g. with a `set\_alarm' intent one would expect a time stamp as its argument. Therefore we think that joint models, or models that treat Intent \& Entity classification together would perform better. We were unable to ascertain this for any of the commercial systems, but Rasa treats them independently (as of Dec 2018). 
(4) None of the platforms use dialogue context for Intent classification and NER - this is another significant limitation, e.g. in understanding elliptical or fragment utterances which depend on the context for their interpretation. 

\VRdel{Dialogflow and Watson support building conversational apps/bots and include a number of prebuilt apps. LUIS is for NLU only; to build a conversation app or a bot their Bot Service is needed. Rasa supports building conversation apps using Rasa Core
While Rasa Core is a machine learning based dialogue framework, the rest of the platforms, viz. Dialogflow, LUIS, and Watson are all rule-based, i.e. one need to manually write rules for dialogue management or dialogue flow.
Dialogflow, LUIS and Watson all have web-based UI for creating and uploading training data, annotating entities and testing the trained model. Rasa NLU is expected to be configured for training and testing at client's servers, and includes a tool
for visualizing and annotating entities.
Each commercial platform is characterised by some restrictions on the maximum number of Intents/Entities/parameters to use etc. Open Source Rasa gives larger flexibility with this.
} 


\vspace*{-\baselineskip}
\begin{table}
    \centering
    \captionsetup{justification=centering}
    \footnotesize
    \begin{tabular}{ | p{1.3cm} | p{5cm} | p{5cm} | }\hline
    Service&Input (Training) &Output (Prediction)\\\hline\hline
    Rasa&JSON or Markdown. Utterances with annotated intents and entities. Can provide synonym and regex features.&JSON. The intent and intent\_ranking with confidence. A list of entities without scores.\\\hline
    Dialogflow&JSON. List of all entity type names and values/synonyms. Utterance samples with annotated intents and entities. Need to specify the expected returning entities as parameters for each intent.&JSON. The intent and entities with values. Overall score returned, not specific to Intent or Entity. Other returned info related to dialogue app.\\\hline
    LUIS&JSON, Phrase list and regex patterns as model features, hierarchical and composites entities. List of all intents and entity type names. Utterance samples with annotated intents and entities&JSON. The intent with confidence. A list of entities with scores\\\hline
    Watson&CSV. List of all utterances with Intent label. List of all Entities with values. No annotated entities in an utterance needed.&JSON. The intent with confidence. A list of entities and confidence for each. Other info related to dialogue app.\\\hline
    \end{tabular}
    \caption{Input Requirements and Output of NLU Services}
    \label{nlu-services-inuputs-outputs}
\end{table}

\vspace*{-\baselineskip}
\vspace*{-\baselineskip}

\section{Data Collection and Annotation}
\label{sec:4}
The evaluation of NLU services was performed in the context of building a SDS, aka Conversational Interface, for a home assistant robot. The home robot is expected to perform a wide variety of tasks, ranging from setting alarms, playing music, search, to movie recommendation, much like existing commercial systems such as Microsoft's Cortana, Apple's Siri, Google Home or Amazon Alexa. Therefore the NLU component in a SDS for such a robot has to understand and be able to respond to a very wide range of user requests and questions, spanning multiple domains, unlike a single domain SDS \VRdel{- most systems -} which only understands and responds to the user in a specific domain. 

\subsection{Data Collection: Crowdsourcing setup}
To build the NLU component we collected real user data via Amazon Mechanical Turk (AMT). 
We designed tasks where the Turker's goal was to answer questions about how people would interact with the home robot, in a wide range of scenarios designed in advance, namely: alarm, audio, audiobook, calendar, cooking, datetime, email, game, general, IoT, lists, music, news, podcasts, general Q\&A, radio, recommendations, social, food takeaway, transport, and weather. 

The questions 
put to Turkers were designed to capture the different requests within each given scenario. In the `calendar' scenario, for example, these pre-designed intents were included: `set\_event', `delete\_event' and `query\_event'. 
An example question for intent `set\_event' is: ``How would you ask your PDA to schedule a meeting with someone?" for which a user's answer example was ``Schedule a chat with Adam on Thursday afternoon".
The Turkers would then type in their answers to these questions and select possible entities from the pre-designed suggested entities list for each of their answers.
The Turkers didn't always follow the instructions fully, e.g. for the specified `delete\_event' Intent, an answer was: ``PDA what is my next event?''; which clearly belongs to `query\_event' Intent. We have manually corrected all such errors either during post-processing or the subsequent annotations.

The data is organized in CSV format which includes information like scenarios, intents, user answers, annotated user answers etc.(See Table \ref{data-anno-snippet} in Appendix). The split training set and test set were converted into different JSON formats for each platform according to the specific requirements of the each platform (see Table \ref{nlu-services-inuputs-outputs}) 

{Our final annotated corpus contains \bf 25716 utterances, annotated for 64 Intents and 54 Entity Types}. 

\subsection{Annotation \& Inter-annotator Agreement}
Since there was a predetermined set of Intents for which we collected data, there was no need for separate Intent annotations(some Intent corrections were needed). We therefore only annotated the data for Entity Tokens \& Entity Types.
Three students were recruited to do the annotations. To calculate inter-annotator agreement, each student annotated the same set of 300 randomly selected utterances. 
Each student then annotated 
a third of the whole dataset, namely, about 8K utterances for annotation.
We used Fleiss's Kappa, suitable for multiple annotators. A match was defined as follows: if there was any overlap between the Entity Tokens (i.e. Partial Tokens Matching), and the annotated Entity Types matched exactly. We achieved moderate agreement ($\kappa = 0.69$) for this task.

\section{Evaluation Experiments}
\label{sec:5}
In this section we describe our evaluation experiments, comparing the performance of the four systems outlined above.
\subsection{Train \& Test Sets} 
Since LUIS caps the size of the training set to 10K, 
 we chose 190 instances of each of the 64 Intents \emph{at random}.  Some of the Intents had slightly fewer instances than 190. This resulted in a {\bf sub-corpus of 11036 utterances} covering all the 64 Intents and 54 Entity Types.
 The Appendix provides more details: Table \ref{data_distr_intents} shows the number of the sentences for each Intent. Table \ref{data_distr_entities} lists the number of entity samples for each Entity Type.
For the evaluation experiments we report below, we performed \textit{10 fold cross-validation} with 90\% of the subcorpus for training and 10\% for testing in each fold.\footnote{
We also note here that our dataset was inevitably unbalanced across the different Intents \& Entities: e.g. some Intents had much fewer instances: \texttt{iot\_wemo} had only 77 instances. But this would affect the performance of the four platforms equally, and thus does not confound the results presented below.}


\subsection{System Versions \& Configurations} Our latest evaluation runs were \VRdel{done at around} completed by the end of March 2018. The service API used was V1.0 for Dialogflow, V2.0 for LUIS. Watson API requests require data as a version parameter which is automatically matched to the closest internal version, where we specified \texttt{2017/04/21}\footnote{At the time of producing the camera-ready version of this paper, we noticed the seemingly recent addition of a `Contextual Entity' annotation tool to Watson, much like e.g. in Rasa. Wed like to stress that this paper does \emph{not} include an evaluation of this feature in Watson NLU}. In our conversational system we run the open source Rasa as our main NLU component because \VRdel{we can have}it allows us to have more control over \VRdel{it for} further developments and extensions. The evaluation done for Rasa was on Version \texttt{0.10.5}, and we used its spacy\_sklearn pipeline which uses Conditional Random Fields for NER and sk-learn (scikit-learn) for Intent classifications. Rasa also provides other built-in components for the processing pipeline, e.g. MITIE, or latest tensorflow\_embedding pipeline.


\section{Results \& Discussion}
\label{sec:6}

We performed 10-fold cross validation for each of the platforms and pairwise t-tests to compare the mean F-scores of every pair of platforms. The results\VRdel{are shown} in Table \ref{overallscores}\VRdel{which shows} show the  micro-average\footnote{Micro-average sums up the individual TP, FP, and FN of all Intent/Entity classes to compute the average metric. 
} scores for Intent and Entity Type classification over 10-fold cross validation. Table \ref{combined_scores} shows the micro-average F-scores of each platform after combining the results of Intents and Entity Types. 
Table \ref{confusion_matrix_intents} and Table \ref{confusion_matrix_entities} in the Appendix show the detailed confusion matrices 
used to calculate the scores of Precision, Recall and F1 for Intents and Entities. 

\vspace*{-\baselineskip} 
\begin{table}[!htb]
    \captionsetup{justification=justified,singlelinecheck=false}
    \begin{minipage}{0.62\linewidth}
     \begin{tabular}{|c|c|c|c|c|c|c|}
     \hline
     \cline{1-4} &  \multicolumn{3}{|c|}{Intent} & \multicolumn{3}{|c|}{Entity} \\
     \hline
     \cline{2-7} & Prec & Rec & F1 & Prec & Rec & F1 \\ \hline
     Rasa & 0.863 & 0.863 & 0.863 & 0.859 & 0.694 & 0.768 \\ \hline
     Dialogflow & 0.870 & 0.859 & 0.864 & 0.782 & 0.709 & 0.743\\ \hline
     LUIS & 0.855 & 0.855 & 0.855 & 0.837 & 0.725 & {\bf0.777}\\ \hline
     Watson & 0.884 & 0.881 & {\bf0.882} & 0.354 & 0.787 & {\bf0.488}\\ \hline
   \end{tabular}
    \caption{Overall Scores for Intent and Entity}
    \label{overallscores}
    \end{minipage}%
    \begin{minipage}{0.6\linewidth}
     \begin{tabular}{|c|c|c|c|}
     \hline
     \cline{2-4} & Prec & Rec & F1 \\
     \hline
     Rasa & 0.862 & 0.787 & 0.822\\
     \hline
     Dialogflow & 0.832 & 0.791 & 0.811\\
     \hline
     LUIS & 0.848 & 0.796 & 0.821\\
     \hline
     Watson & 0.540 & 0.838 & {\bf0.657}\\
     \hline
   \end{tabular}
      \caption{Combined Overall Scores}
      \label{combined_scores}
    \end{minipage} 
\end{table}
\vspace*{-\baselineskip}



Performing significance tests on separate Intent and Entity scores in Table \ref{overallscores} revealed:
For Intent, there is no significant difference between Dialogflow, LUIS and Rasa. Watson F1 score (0.882) is significantly higher than other platforms ($p<0.05$, with large or very large effects sizes - Cohen's D). 
However, for Entities, Watson achieves significantly 
lower F1 scores ($p<0.05$, with large or very large effects sizes - Cohen's D) due to its very low Precision. One explanation for this is the high number of Entity candidates produced in its predictions, leading to a high number of False Positives\footnote{Interestingly, Watson only requires a list of possible entities rather than entity annotation in utterances as other platforms do (See Table \ref{nlu-services-inuputs-outputs})}. It also shows that there are significant differences for Entity F1 score between Dialogflow, 
 LUIS 
 and Rasa. 
 LUIS achieved the top F1 score (0.777) on Entities.




Table \ref{combined_scores} shows that \VRdel{they} all NLU services have quite close F1 scores except for Watson which had significantly 
lower score ($p<0.05$, with large or very large effects sizes - Cohen's D) due to its lower entity score as discussed above. The significance test shows no significant differences between Dialogflow, LUIS and Rasa.

The detailed data analysis results in the Appendix (see Table \ref{data_distr_intents} and Table \ref{data_distr_entities}) for fold-1\footnote{Tables for other folds are omitted for space reason} reveal that distributions of Intents and Entities are imbalanced in the datasets. 
Also, our data contains some noisy Entity annotations, often caused by ambiguities, which our simplified annotation scheme was not able to capture. For example, an utterance in the pattern ``play xxx please" where xxx could be any entity from song\_name, audiobook\_name, radio\_name, posdcasts\_name or game\_name, e.g. ``play space invaders please" which could be annotated the entity as [song\_name : space invaders] or [game\_name : space invaders]. 
This type of Intent ambiguity that can only be resolved by more sophisticated approaches that incorporate domain knowledge and the dialogue context. 
Nevertheless, despite the noisiness of the data, we believe that it represents a real-world use case for NLU engines.

\VRdel{We see our current version of the training/test datasets contain some unbalanced data samples data (Refer to Table \ref{data_distr_intents} and Table \ref{data_distr_entities} in the Appendix) and contain some ambiguous samples in the annotated data especially in the Entity annotations as our annotation was by purpose designed to be simplified and fast for quick prototype, thus we did not aim to address inherent ambiguity that can be only resolved by more sophisticated approaches like dependency parsing / logical forms and combing with the domain knowledge. We think they are fair enough in terms of comparing different NLU systems as they are all fed into exact same training data and testing data, but for the system deployment purposes, the datasets need to be improved. }

\section{Conclusion}

The contributions of this paper are two-fold: First, we present and release a large NLU dataset in the context of a real-world use case of a home robot, covering 21 domains with 64 Intents and 54 Entity Types.
Secondly, we perform a comparative evaluation on this data of some of the most popular NLU services -- namely the commercial platforms Dialogflow, LUIS, Watson and the open source Rasa.

The results show they all have similar functions/features and achieve similar performance in terms of combined F-scores. However, when dividing out results for Intent and Entity Type recognition, we find that Watson has significant higher F-scores for Intent, but significantly lower scores for Entity Type. This was due to its high number of false positives produced in its Entity predictions. As noted earlier, we have \emph{not} here evaluated Watson's recent `Contexual Entity' annotation tool.


In future work, we hope to continuously improve the data quality and observe its impact on NLU performance. However, we do believe that noisy data presents an interesting real-world use-case for testing current NLU services.
We are also working on extending the data set with spoken user utterances, rather than typed input. This will allow us to investigate the impact of ASR errors on NLU performance.

\newpage

\section*{Appendix}\label{appendix-a}
We provide some examples of the data annotation and the training inputs to each of the 4 platforms in Table \ref{data-anno-snippet}, Listing \ref{rasa-traindata-snippet}, \ref{luis-traindata-snippet}, \ref{watson-traindata-snippet} and \ref{dialogflow-traindata-snippet}.

We also provide more details on the 
train and test data distribution, as well as the Confusion Matrix for the first fold (Fold\_1) of the 10-Fold Cross Validation. 
Table \ref{data_distr_intents} shows the number of the sentences for each Intent in each dataset. Table \ref{data_distr_entities} lists the number of entity samples for each Entity Type in each dataset. Table \ref{confusion_matrix_intents} and Table \ref{confusion_matrix_entities} show the confusion matrices 
used to calculate the scores of Precision, Recall and F1 for Intents and Entities. The TP, FP, FN and TN in the tables are short for True Positive, False Positive, False Negative and True Negative respectively.
\vspace*{-\baselineskip} 
%
\begin{table}[ht]
\scriptsize
\begin{tabular}{|l|l|l|l|l|}\hline
userid & answerid & scenario & intent &  answer\_annotation \\\hline\hline

1 & 2 & alarm & set & wake me up at [time : nine am] on [date : friday] \\\hline
2 & 558 & alarm & remove & cancel my [time : seven am] alarm\\\hline
2 & 559 & alarm & remove & remove the alarm set for [time : ten pm]\\\hline
2 & 561 & alarm & query & what alarms i have set\\\hline
502 & 12925 & calendar & query & what is the time for [event\_name : jimmy's party]\\\hline
653 & 17462 & calendar & query & what is up in my schedule [date : today]\\\hline
2 & 564 & calendar & remove & please cancel all my events for [date : today]\\\hline
2 & 586 & play & music & i'd like to hear [artist\_name : queen's] [song\_name : barcelona]\\\hline
65 & 2813 & play & radio & play a [radio\_name : pop station] on the radio\\\hline
740 & 19087 & play & podcasts & play my favorite podcast\\\hline
1 & 1964 & weather & query & tell me the weather in [place\_name : barcelona] in [time : two days from now]\\\hline
92 & 3483 & weather & query & what is the current [weather\_descriptor : temperature] outside\\\hline
394 & 10448 & email & sendemail & send an email to [person : sarah] about [event\_name : brunch] [date : today]\\\hline
4 & 649 & email & query & has the [business\_name : university of greenwich] emailed me\\\hline
2 & 624 & takeaway & order & please order some [food\_type : sushi] for [meal\_type : dinner]\\\hline
38 & 2045 & takeaway & query & search if the [business\_type : restaurant] does [order\_type : take out]\\\hline

\end{tabular}
\caption{Data annotation example snippet}\label{data-anno-snippet}
\end{table}

\vspace*{-\baselineskip} 
\vspace*{-\baselineskip} 

\begin{lstlisting}[caption={Rasa train data example snippet}, language=json,firstnumber=1, label=rasa-traindata-snippet]
{
  "rasa_nlu_data": {
    "common_examples": [  {
        "text": "lower the lights in the bedroom",
        "intent": "iot_hue_lightdim",
        "entities": [   {
            "start": 24,
            "end": 31,
            "value": "bedroom",
            "entity": "house_place"
          } ]  },
      {
        "text": "dim the lights in my bedroom",
        "intent": "iot_hue_lightdim",
        "entities": [  {
            "start": 21,
            "end": 28,
            "value": "bedroom",
            "entity": "house_place"
          } ]   },
       ... ...
       ]
}
\end{lstlisting}

\begin{lstlisting}[caption={LUIS train data example snippet}, language=json,firstnumber=1, label=luis-traindata-snippet]
{
  "intents": [   
    {  "name": "play_podcasts"   },
    {  "name": "music_query"     },
    .......
  ],
  "entities": [  {
      "name": "Hier2",
      "children": [
        "business_type",  "event_name",  "place_name",  "time", "timeofday" ]  },
    ... ...
  ],
  "utterances": [  {
      "text": "call a taxi for me",
      "intent": "transport_taxi",
      "entities": [ {
          "startPos": 7,
          "endPos": 10,
          "value": "taxi",
          "entity": "Hier9::transport_type"
        }    ]   }, 
     ... ...    
    ]
}
\end{lstlisting}

\begin{lstlisting}[caption={Watson train data example snippet}, language=json,firstnumber=1, label=watson-traindata-snippet]
---- Watson Entity list ----

joke_type,nice  joke_type,funny joke_type,sarcastic
 ... ... 
relation,mum    relation,dad    person,ted
... ... 
person,emma     person,bina     person,daniel bell

---- Watson utterance and Intent list ----

give me the weather for merced at three pm, weather_query
weather this week,weather_query
find weather report,weather_query
should i wear a hat today,weather_query
what should i wear is it cold outside,weather_query
is it going to snow tonight,weather_query
\end{lstlisting}


\begin{lstlisting}[caption={Dialogflow train data example snippet}, language=json,firstnumber=1, label=dialogflow-traindata-snippet]
---- Dialogflow Entity list ----
{
  "id": "... ...",
  "name": "artist_name",
  "isOverridable": true,
  "entries": [   {
      "value": "aaron carter",
      "synonyms": [
        "aaron carter"
      ] },
    {
      "value": "adele",
      "synonyms": [ "adele"   ]
    }  ],
  "isEnum": false,
  "automatedExpansion": true
}

---- Dialogflow "alarm_query" Intent annotation ----
{
  "userSays": [ {
      "id": " ... ... ",
      "data": [ {  "text": "checkout "   },
        {
          "text": "today",
          "alias": "date",
          "meta": "@date",
          "userDefined": true
        },
        {  "text": " alarm of meeting"   }
      ],
      "isTemplate": false,
      "count": 0
    },
   ... ...
] }
\end{lstlisting}

\begin{table}[ht]
\centering
\scriptsize
\begin{tabular}{|c|c|c|c|c|c|c|c|c|c|c|c|}\hline

    Intent & Total &  Train & Test & Intent & Total &  Train & Test & Intent & Total &  Train & Test\\\hline
    
    alarm\_query & 194 & 175 & 19 & general\_negate & 194 & 175 & 19 & play\_music & 194 & 175 & 19\\\hline
    alarm\_remove & 117 & 106 & 11 & general\_praise & 194 & 175 & 19 & play\_podcasts & 194 & 175 & 1\\\hline
    alarm\_set & 194 & 175 & 19 & general\_quirky & 194 & 175 & 19 & play\_radio & 194 & 175 & 19\\\hline
    audio\_volume\_down & 80 & 72 & 8 & general\_repeat & 194 & 175 & 19 & qa\_currency & 194 & 175 & 19\\\hline
    audio\_volume\_mute & 157 & 142 & 15 & iot\_cleaning & 167 & 151 & 16 & qa\_definition & 194 & 175 & 19\\\hline
    audio\_volume\_up & 139 & 126 & 13 & iot\_coffee & 194 & 175 & 19 & qa\_factoid & 194 & 175 & 19\\\hline
    calendar\_query & 194 & 175 & 19 & iot\_hue\_lightchange & 194 & 175 & 19 & qa\_maths & 148 & 134 & 14\\\hline
    calendar\_remove & 194 & 175 & 19 & iot\_hue\_lightdim & 126 & 114 & 12 & qa\_stock & 194 & 175 & 19\\\hline
    calendar\_set & 194 & 175 & 19 & iot\_hue\_lightoff & 194 & 175 & 19 & rec\_events & 194 & 175 & 19\\\hline
    cooking\_recipe & 194 & 175 & 19 & iot\_hue\_lighton & 38 & 35 & 3 & rec\_locations & 194 & 175 & 19\\\hline
    datetime\_convert & 87 & 79 & 8 & iot\_hue\_lightup & 140 & 126 & 14 & rec\_movies & 107 & 97 & 10\\\hline
    datetime\_query & 194 & 175 & 19 & iot\_wemo\_off & 98 & 89 & 9 & social\_post & 194 & 175 & 19\\\hline
    email\_addcontact & 87 & 79 & 8 & iot\_wemo\_on & 76 & 69 & 7 & social\_query & 183 & 165 & 18\\\hline
    email\_query & 194 & 175 & 19 & lists\_createoradd & 194 & 175 & 19 & takeaway\_order & 194 & 175 & 19\\\hline
    email\_querycontact & 194 & 175 & 19 & lists\_query & 194 & 175 & 19 & takeaway\_query & 194 & 175 & 19\\\hline
    email\_sendemail & 194 & 175 & 19 & lists\_remove & 194 & 175 & 19 & transport\_query & 194 & 175 & 19\\\hline
    general\_affirm & 194 & 175 & 19 & music\_likeness & 180 & 162 & 18 & transport\_taxi & 181 & 163 & 18\\\hline
    general\_commandstop & 194 & 175 & 19 & music\_query & 194 & 175 & 19 & transport\_ticket & 194 & 175 & 19\\\hline
    general\_confirm & 194 & 175 & 19 & music\_settings & 77 & 70 & 7 & transport\_traffic & 190 & 171 & 19\\\hline
    general\_dontcare & 194 & 175 & 19 & news\_query & 194 & 175 & 19 & weather\_query & 194 & 175 & 19\\\hline
    general\_explain & 194 & 175 & 19 & play\_audiobook & 194 & 175 & 19 &  &  &  & \\\hline
    general\_joke & 122 & 110 & 12 & play\_game & 194 & 175 & 19 &  &  &  & \\\hline

\end{tabular}
\caption{Data Distribution for Intents in Fold\_1}
\label{data_distr_intents}
\end{table}


\begin{table}[ht]
\centering
\scriptsize
\begin{tabular}{|c|c|c|c|c|c|c|c|c|}\hline

    Entity & Trainset & Testset & Entity & Trainset & Testset & Entity & Trainset & Testset\\\hline\hline
    
    alarm\_type & 14 & 0 & event\_name & 352 & 48 & person & 468 & 42\\\hline
    app\_name & 32 & 5 & food\_type & 302 & 25 & personal\_info & 100 & 14\\\hline
    artist\_name & 91 & 11 & game\_name & 133 & 17 & place\_name & 869 & 95\\\hline
    audiobook\_author & 10 & 1 & game\_type & 1 & 0 & player\_setting & 190 & 19\\\hline
    audiobook\_name & 97 & 10 & general\_frequency & 27 & 5 & playlist\_name & 22 & 1\\\hline
    business\_name & 394 & 41 & house\_place & 259 & 25 & podcast\_descriptor & 67 & 6\\\hline
    business\_type & 199 & 19 & ingredient & 17 & 4 & podcast\_name & 44 & 2\\\hline
    change\_amount & 57 & 9 & joke\_type & 59 & 4 & radio\_name & 99 & 12\\\hline
    coffee\_type & 31 & 4 & list\_name & 211 & 13 & relation & 127 & 13\\\hline
    color\_type & 135 & 11 & meal\_type & 37 & 0 & song\_name & 51 & 9\\\hline
    cooking\_type & 10 & 0 & media\_type & 370 & 40 & time & 511 & 62\\\hline
    currency\_name & 296 & 35 & movie\_name & 18 & 0 & time\_zone & 59 & 7\\\hline
    date & 905 & 85 & movie\_type & 13 & 0 & timeofday & 150 & 26\\\hline
    definition\_word & 158 & 16 & music\_album & 1 & 0 & transport\_agency & 59 & 10\\\hline
    device\_type & 353 & 41 & music\_descriptor & 17 & 2 & transport\_descriptor & 11 & 0\\\hline
    drink\_type & 6 & 0 & music\_genre & 72 & 8 & transport\_name & 10 & 2\\\hline
    email\_address & 38 & 5 & news\_topic & 75 & 9 & transport\_type & 363 & 35\\\hline
    email\_folder & 17 & 1 & order\_type & 151 & 17 & weather\_descriptor & 95 & 14\\\hline

\end{tabular}
\caption{Data Distribution for Entities in Fold\_1}\label{data_distr_entities}
\end{table}



\begin{table}[ht]
\centering
\scriptsize

     \begin{tabular}{|c|c|c|c|c|c|c|c|c|c|c|c|c|c|c|c|c|}  \hline
     \cline{1-5} & \multicolumn{4}{|c|}{Rasa} & \multicolumn{4}{|c|}{Dialogflow} & \multicolumn{4}{|c|}{LUIS} & \multicolumn{4}{|c|}{Watson}\\\hline
     \cline{1-17} Intent & TP & FP & FN & TN  &  TP & FP & FN & TN  &  TP & FP & FN & TN  &  TP & FP & FN & TN \\\hline
		alarm\_query & 17 & 1 & 2 & 1056 & 19 & 0 & 0 & 1057 & 18 & 2 & 1 & 1055 & 19 & 0 & 0 & 1057\\\hline
		alarm\_remove & 11 & 0 & 0 & 1065 & 10 & 2 & 1 & 1063 & 9 & 0 & 2 & 1065 & 11 & 0 & 0 & 1065\\\hline
		alarm\_set & 18 & 3 & 1 & 1054 & 17 & 4 & 2 & 1053 & 17 & 3 & 2 & 1054 & 17 & 3 & 2 & 1054\\\hline
		audio\_volume\_down & 7 & 1 & 1 & 1067 & 8 & 0 & 0 & 1068 & 7 & 0 & 1 & 1068 & 8 & 0 & 0 & 1068\\\hline
		audio\_volume\_mute & 13 & 1 & 2 & 1060 & 14 & 0 & 1 & 1061 & 12 & 1 & 3 & 1060 & 14 & 1 & 1 & 1060\\\hline
		audio\_volume\_up & 12 & 3 & 1 & 1060 & 13 & 0 & 0 & 1063 & 12 & 3 & 1 & 1060 & 12 & 3 & 1 & 1060\\\hline
		calendar\_query & 11 & 10 & 8 & 1047 & 13 & 18 & 6 & 1039 & 11 & 6 & 8 & 1051 & 10 & 8 & 9 & 1049\\\hline
		calendar\_remove & 17 & 0 & 2 & 1057 & 18 & 1 & 1 & 1056 & 18 & 2 & 1 & 1055 & 19 & 1 & 0 & 1056\\\hline
		calendar\_set & 16 & 2 & 3 & 1055 & 14 & 2 & 5 & 1055 & 14 & 4 & 5 & 1053 & 16 & 3 & 3 & 1054\\\hline
		cooking\_recipe & 15 & 1 & 4 & 1056 & 11 & 2 & 8 & 1055 & 13 & 4 & 6 & 1053 & 15 & 1 & 4 & 1056\\\hline
		datetime\_convert & 5 & 2 & 3 & 1066 & 7 & 4 & 1 & 1064 & 7 & 2 & 1 & 1066 & 8 & 2 & 0 & 1066\\\hline
		datetime\_query & 17 & 4 & 2 & 1053 & 18 & 9 & 1 & 1048 & 17 & 4 & 2 & 1053 & 18 & 4 & 1 & 1053\\\hline
		email\_addcontact & 8 & 3 & 0 & 1065 & 8 & 0 & 0 & 1068 & 8 & 0 & 0 & 1068 & 8 & 2 & 0 & 1066\\\hline
		email\_query & 17 & 1 & 2 & 1056 & 18 & 1 & 1 & 1056 & 15 & 3 & 4 & 1054 & 17 & 2 & 2 & 1055\\\hline
		email\_querycontact & 11 & 4 & 8 & 1053 & 13 & 3 & 6 & 1054 & 14 & 4 & 5 & 1053 & 14 & 3 & 5 & 1054\\\hline
		email\_sendemail & 17 & 1 & 2 & 1056 & 16 & 1 & 3 & 1056 & 16 & 4 & 3 & 1053 & 17 & 2 & 2 & 1055\\\hline
		general\_affirm & 19 & 1 & 0 & 1056 & 19 & 0 & 0 & 1057 & 19 & 0 & 0 & 1057 & 19 & 1 & 0 & 1056\\\hline
		general\_commandstop & 19 & 0 & 0 & 1057 & 18 & 1 & 1 & 1056 & 19 & 0 & 0 & 1057 & 19 & 1 & 0 & 1056\\\hline
		general\_confirm & 19 & 1 & 0 & 1056 & 19 & 0 & 0 & 1057 & 19 & 0 & 0 & 1057 & 19 & 0 & 0 & 1057\\\hline
		general\_dontcare & 19 & 0 & 0 & 1057 & 19 & 1 & 0 & 1056 & 18 & 1 & 1 & 1056 & 19 & 2 & 0 & 1055\\\hline
		general\_explain & 19 & 1 & 0 & 1056 & 19 & 0 & 0 & 1057 & 18 & 0 & 1 & 1057 & 19 & 2 & 0 & 1055\\\hline
		general\_joke & 11 & 0 & 1 & 1064 & 12 & 0 & 0 & 1064 & 12 & 0 & 0 & 1064 & 12 & 0 & 0 & 1064\\\hline
		general\_negate & 18 & 0 & 1 & 1057 & 19 & 0 & 0 & 1057 & 19 & 1 & 0 & 1056 & 19 & 0 & 0 & 1057\\\hline
		general\_praise & 18 & 1 & 1 & 1056 & 19 & 0 & 0 & 1057 & 19 & 1 & 0 & 1056 & 18 & 1 & 1 & 1056\\\hline
		general\_quirky & 11 & 22 & 8 & 1035 & 4 & 2 & 15 & 1055 & 8 & 16 & 11 & 1041 & 7 & 9 & 12 & 1048\\\hline
		general\_repeat & 19 & 0 & 0 & 1057 & 19 & 1 & 0 & 1056 & 19 & 0 & 0 & 1057 & 19 & 0 & 0 & 1057\\\hline
		iot\_cleaning & 14 & 1 & 2 & 1059 & 13 & 6 & 3 & 1054 & 16 & 1 & 0 & 1059 & 16 & 1 & 0 & 1059\\\hline
		iot\_coffee & 18 & 3 & 1 & 1054 & 18 & 1 & 1 & 1056 & 18 & 0 & 1 & 1057 & 19 & 1 & 0 & 1056\\\hline
		iot\_hue\_lightchange & 15 & 1 & 4 & 1056 & 14 & 3 & 5 & 1054 & 15 & 4 & 4 & 1053 & 13 & 3 & 6 & 1054\\\hline
		iot\_hue\_lightdim & 12 & 0 & 0 & 1064 & 11 & 0 & 1 & 1064 & 10 & 1 & 2 & 1063 & 11 & 1 & 1 & 1063\\\hline
		iot\_hue\_lightoff & 17 & 2 & 2 & 1055 & 15 & 0 & 4 & 1057 & 17 & 1 & 2 & 1056 & 17 & 2 & 2 & 1055\\\hline
		iot\_hue\_lighton & 3 & 3 & 0 & 1070 & 3 & 3 & 0 & 1070 & 2 & 3 & 1 & 1070 & 3 & 3 & 0 & 1070\\\hline
		iot\_hue\_lightup & 9 & 1 & 5 & 1061 & 11 & 1 & 3 & 1061 & 11 & 0 & 3 & 1062 & 11 & 2 & 3 & 1060\\\hline
		iot\_wemo\_off & 9 & 2 & 0 & 1065 & 8 & 4 & 1 & 1063 & 9 & 4 & 0 & 1063 & 9 & 2 & 0 & 1065\\\hline
		iot\_wemo\_on & 5 & 2 & 2 & 1067 & 5 & 1 & 2 & 1068 & 4 & 3 & 3 & 1066 & 6 & 1 & 1 & 1068\\\hline
		lists\_createoradd & 16 & 2 & 3 & 1055 & 16 & 6 & 3 & 1051 & 16 & 5 & 3 & 1052 & 18 & 3 & 1 & 1054\\\hline
		lists\_query & 16 & 3 & 3 & 1054 & 16 & 5 & 3 & 1052 & 16 & 3 & 3 & 1054 & 14 & 2 & 5 & 1055\\\hline
		lists\_remove & 17 & 1 & 2 & 1056 & 18 & 3 & 1 & 1054 & 18 & 2 & 1 & 1055 & 18 & 0 & 1 & 1057\\\hline
		music\_likeness & 12 & 4 & 6 & 1054 & 13 & 5 & 5 & 1053 & 13 & 3 & 5 & 1055 & 14 & 1 & 4 & 1057\\\hline
		music\_query & 13 & 0 & 6 & 1057 & 11 & 3 & 8 & 1054 & 10 & 4 & 9 & 1053 & 11 & 2 & 8 & 1055\\\hline
		music\_settings & 6 & 2 & 1 & 1067 & 4 & 2 & 3 & 1067 & 7 & 0 & 0 & 1069 & 7 & 2 & 0 & 1067\\\hline
		news\_query & 13 & 9 & 6 & 1048 & 10 & 4 & 9 & 1053 & 13 & 3 & 6 & 1054 & 14 & 1 & 5 & 1056\\\hline
		play\_audiobook & 16 & 3 & 3 & 1054 & 13 & 8 & 6 & 1049 & 17 & 1 & 2 & 1056 & 16 & 2 & 3 & 1055\\\hline
		play\_game & 15 & 5 & 4 & 1052 & 13 & 2 & 6 & 1055 & 13 & 2 & 6 & 1055 & 13 & 2 & 6 & 1055\\\hline
		play\_music & 13 & 4 & 6 & 1053 & 16 & 5 & 3 & 1052 & 12 & 11 & 7 & 1046 & 12 & 14 & 7 & 1043\\\hline
		play\_podcasts & 17 & 0 & 2 & 1057 & 14 & 1 & 5 & 1056 & 16 & 0 & 3 & 1057 & 17 & 1 & 2 & 1056\\\hline
		play\_radio & 15 & 1 & 4 & 1056 & 15 & 2 & 4 & 1055 & 17 & 1 & 2 & 1056 & 15 & 2 & 4 & 1055\\\hline
		qa\_currency & 17 & 1 & 2 & 1056 & 16 & 0 & 3 & 1057 & 18 & 0 & 1 & 1057 & 18 & 0 & 1 & 1057\\\hline
		qa\_definition & 19 & 0 & 0 & 1057 & 13 & 2 & 6 & 1055 & 18 & 0 & 1 & 1057 & 18 & 1 & 1 & 1056\\\hline
		qa\_factoid & 10 & 13 & 9 & 1044 & 7 & 9 & 12 & 1048 & 15 & 15 & 4 & 1042 & 14 & 8 & 5 & 1049\\\hline
		qa\_maths & 14 & 2 & 0 & 1060 & 12 & 2 & 2 & 1060 & 13 & 4 & 1 & 1058 & 14 & 1 & 0 & 1061\\\hline
		qa\_stock & 19 & 2 & 0 & 1055 & 19 & 1 & 0 & 1056 & 19 & 0 & 0 & 1057 & 19 & 1 & 0 & 1056\\\hline
		recommendation\_events & 13 & 2 & 6 & 1055 & 14 & 6 & 5 & 1051 & 16 & 3 & 3 & 1054 & 15 & 2 & 4 & 1055\\\hline
		recommendation\_locations & 16 & 1 & 3 & 1056 & 15 & 1 & 4 & 1056 & 17 & 2 & 2 & 1055 & 16 & 1 & 3 & 1056\\\hline
		recommendation\_movies & 8 & 2 & 2 & 1064 & 8 & 2 & 2 & 1064 & 9 & 1 & 1 & 1065 & 10 & 2 & 0 & 1064\\\hline
		social\_post & 18 & 3 & 1 & 1054 & 17 & 4 & 2 & 1053 & 18 & 1 & 1 & 1056 & 19 & 1 & 0 & 1056\\\hline
		social\_query & 16 & 5 & 2 & 1053 & 14 & 8 & 4 & 1050 & 17 & 3 & 1 & 1055 & 17 & 3 & 1 & 1055\\\hline
		takeaway\_order & 12 & 0 & 7 & 1057 & 16 & 2 & 3 & 1055 & 16 & 4 & 3 & 1053 & 16 & 1 & 3 & 1056\\\hline
		takeaway\_query & 18 & 6 & 1 & 1051 & 19 & 3 & 0 & 1054 & 16 & 2 & 3 & 1055 & 18 & 3 & 1 & 1054\\\hline
		transport\_query & 16 & 3 & 3 & 1054 & 17 & 3 & 2 & 1054 & 13 & 3 & 6 & 1054 & 14 & 5 & 5 & 1052\\\hline
		transport\_taxi & 17 & 2 & 1 & 1056 & 17 & 1 & 1 & 1057 & 18 & 0 & 0 & 1058 & 18 & 1 & 0 & 1057\\\hline
		transport\_ticket & 16 & 1 & 3 & 1056 & 17 & 0 & 2 & 1057 & 16 & 1 & 3 & 1056 & 16 & 2 & 3 & 1055\\\hline
		transport\_traffic & 18 & 1 & 1 & 1056 & 18 & 1 & 1 & 1056 & 18 & 1 & 1 & 1056 & 19 & 2 & 0 & 1055\\\hline
		weather\_query & 16 & 2 & 3 & 1055 & 12 & 2 & 7 & 1055 & 13 & 5 & 6 & 1052 & 13 & 2 & 6 & 1055\\\hline

\end{tabular}
\caption{Confusion Matrix summary for Intents in Fold\_1}\label{confusion_matrix_intents}
\end{table}



\begin{table}[ht]
\centering
\scriptsize
     \begin{tabular}{|c|c|c|c|c|c|c|c|c|c|c|c|c|c|c|c|c|}  \hline
     \cline{1-5} &  \multicolumn{4}{|c|}{Rasa} & \multicolumn{4}{|c|}{Dialogflow} & \multicolumn{4}{|c|}{LUIS}  & \multicolumn{4}{|c|}{Watson}\\\hline
     \cline{1-17} Entity & TP & FP & FN & TN & TP & FP & FN & TN & TP & FP & FN & TN & TP & FP & FN & TN \\\hline
		app\_name & 3 & 0 & 2 & 1071 & 2 & 1 & 3 & 1070 & 3 & 0 & 2 & 1071 & 4 & 10 & 1 & 1061\\\hline
		artist\_name & 3 & 0 & 8 & 1065 & 5 & 1 & 6 & 1064 & 4 & 2 & 7 & 1063 & 3 & 1 & 8 & 1064\\\hline
		audiobook\_author & 0 & 0 & 1 & 1075 & 0 & 0 & 1 & 1075 & 0 & 0 & 1 & 1075 & 0 & 0 & 1 & 1075\\\hline
		audiobook\_name & 2 & 3 & 8 & 1063 & 6 & 2 & 4 & 1064 & 5 & 1 & 5 & 1065 & 6 & 3 & 4 & 1063\\\hline
		business\_name & 25 & 12 & 16 & 1027 & 32 & 8 & 9 & 1029 & 32 & 5 & 9 & 1031 & 29 & 30 & 12 & 1008\\\hline
		business\_type & 15 & 2 & 4 & 1055 & 13 & 1 & 6 & 1056 & 14 & 5 & 5 & 1054 & 16 & 45 & 3 & 1014\\\hline
		change\_amount & 7 & 0 & 2 & 1067 & 6 & 2 & 3 & 1065 & 8 & 2 & 1 & 1065 & 6 & 12 & 3 & 1056\\\hline
		coffee\_type & 1 & 0 & 3 & 1072 & 2 & 1 & 2 & 1071 & 2 & 0 & 2 & 1072 & 2 & 4 & 2 & 1068\\\hline
		color\_type & 8 & 2 & 3 & 1063 & 8 & 1 & 3 & 1064 & 8 & 1 & 3 & 1064 & 9 & 26 & 2 & 1042\\\hline
		currency\_name & 25 & 0 & 10 & 1058 & 14 & 0 & 21 & 1058 & 28 & 4 & 7 & 1056 & 31 & 12 & 4 & 1049\\\hline
		date & 77 & 8 & 8 & 983 & 74 & 25 & 11 & 969 & 78 & 9 & 7 & 984 & 80 & 30 & 5 & 971\\\hline
		definition\_word & 7 & 2 & 9 & 1058 & 10 & 3 & 6 & 1057 & 11 & 4 & 5 & 1056 & 6 & 104 & 10 & 961\\\hline
		device\_type & 33 & 0 & 8 & 1035 & 24 & 10 & 17 & 1027 & 33 & 6 & 8 & 1029 & 38 & 76 & 3 & 963\\\hline
		email\_address & 4 & 0 & 1 & 1071 & 4 & 1 & 1 & 1070 & 3 & 2 & 2 & 1071 & 1 & 0 & 4 & 1071\\\hline
		email\_folder & 1 & 0 & 0 & 1075 & 1 & 0 & 0 & 1075 & 1 & 0 & 0 & 1075 & 1 & 0 & 0 & 1075\\\hline
		event\_name & 27 & 4 & 21 & 1024 & 25 & 25 & 23 & 1005 & 24 & 6 & 24 & 1023 & 30 & 56 & 18 & 973\\\hline
		food\_type & 13 & 3 & 12 & 1048 & 16 & 5 & 9 & 1046 & 16 & 4 & 9 & 1047 & 17 & 16 & 8 & 1040\\\hline
		game\_name & 7 & 2 & 10 & 1057 & 11 & 2 & 6 & 1057 & 12 & 0 & 5 & 1059 & 9 & 2 & 8 & 1057\\\hline
		general\_frequency & 1 & 1 & 4 & 1070 & 0 & 0 & 5 & 1071 & 2 & 0 & 3 & 1071 & 3 & 3 & 2 & 1069\\\hline
		house\_place & 22 & 1 & 3 & 1050 & 22 & 10 & 3 & 1042 & 24 & 1 & 1 & 1050 & 25 & 18 & 0 & 1033\\\hline
		ingredient & 0 & 0 & 4 & 1072 & 1 & 0 & 3 & 1072 & 0 & 1 & 4 & 1072 & 1 & 3 & 3 & 1069\\\hline
		joke\_type & 3 & 1 & 1 & 1071 & 3 & 0 & 1 & 1072 & 3 & 2 & 1 & 1070 & 2 & 53 & 2 & 1019\\\hline
		list\_name & 9 & 7 & 4 & 1056 & 6 & 2 & 7 & 1061 & 10 & 5 & 3 & 1058 & 7 & 56 & 6 & 1010\\\hline
		media\_type & 29 & 4 & 11 & 1033 & 26 & 24 & 14 & 1013 & 31 & 11 & 9 & 1026 & 34 & 81 & 6 & 961\\\hline
		music\_descriptor & 0 & 0 & 2 & 1074 & 0 & 0 & 2 & 1074 & 0 & 0 & 2 & 1074 & 0 & 4 & 2 & 1070\\\hline
		music\_genre & 6 & 1 & 2 & 1067 & 7 & 2 & 1 & 1066 & 6 & 1 & 2 & 1067 & 7 & 8 & 1 & 1060\\\hline
		news\_topic & 0 & 2 & 9 & 1065 & 3 & 3 & 6 & 1064 & 2 & 4 & 7 & 1063 & 3 & 18 & 6 & 1049\\\hline
		order\_type & 14 & 3 & 3 & 1056 & 12 & 3 & 5 & 1056 & 13 & 2 & 4 & 1057 & 17 & 8 & 0 & 1051\\\hline
		person & 31 & 14 & 11 & 1021 & 31 & 12 & 11 & 1023 & 30 & 7 & 12 & 1028 & 27 & 36 & 15 & 999\\\hline
		personal\_info & 5 & 0 & 9 & 1063 & 5 & 1 & 9 & 1062 & 7 & 4 & 7 & 1059 & 12 & 58 & 2 & 1011\\\hline
		place\_name & 65 & 22 & 30 & 971 & 66 & 17 & 29 & 976 & 71 & 5 & 24 & 986 & 76 & 39 & 19 & 961\\\hline
		player\_setting & 13 & 2 & 6 & 1056 & 9 & 3 & 10 & 1055 & 16 & 7 & 3 & 1052 & 18 & 71 & 1 & 988\\\hline
		playlist\_name & 0 & 0 & 1 & 1075 & 0 & 0 & 1 & 1075 & 0 & 0 & 1 & 1075 & 0 & 0 & 1 & 1075\\\hline
		podcast\_descriptor & 5 & 1 & 1 & 1069 & 4 & 1 & 2 & 1069 & 5 & 2 & 1 & 1068 & 5 & 9 & 1 & 1061\\\hline
		podcast\_name & 0 & 0 & 2 & 1074 & 0 & 0 & 2 & 1074 & 1 & 2 & 1 & 1072 & 0 & 111 & 2 & 968\\\hline
		radio\_name & 4 & 2 & 8 & 1063 & 6 & 2 & 6 & 1062 & 7 & 5 & 5 & 1060 & 2 & 17 & 10 & 1048\\\hline
		relation & 8 & 0 & 5 & 1063 & 6 & 4 & 7 & 1059 & 7 & 1 & 6 & 1063 & 10 & 4 & 3 & 1059\\\hline
		song\_name & 4 & 1 & 5 & 1066 & 5 & 2 & 4 & 1065 & 3 & 1 & 6 & 1066 & 3 & 13 & 6 & 1055\\\hline
		time & 53 & 3 & 9 & 1013 & 45 & 18 & 17 & 1002 & 49 & 12 & 13 & 1010 & 55 & 119 & 7 & 928\\\hline
		time\_zone & 2 & 0 & 5 & 1071 & 3 & 1 & 4 & 1070 & 2 & 1 & 5 & 1070 & 6 & 63 & 1 & 1019\\\hline
		timeofday & 23 & 3 & 3 & 1047 & 13 & 3 & 13 & 1047 & 22 & 4 & 4 & 1047 & 26 & 4 & 0 & 1046\\\hline
		transport\_agency & 10 & 0 & 0 & 1066 & 10 & 0 & 0 & 1066 & 10 & 0 & 0 & 1066 & 10 & 0 & 0 & 1066\\\hline
		transport\_name & 0 & 0 & 2 & 1074 & 0 & 0 & 2 & 1074 & 0 & 0 & 2 & 1074 & 0 & 0 & 2 & 1074\\\hline
		transport\_type & 35 & 1 & 0 & 1040 & 14 & 1 & 21 & 1041 & 34 & 4 & 1 & 1039 & 35 & 7 & 0 & 1035\\\hline
		weather\_descriptor & 5 & 1 & 9 & 1063 & 7 & 3 & 7 & 1061 & 7 & 2 & 7 & 1062 & 8 & 12 & 6 & 1053\\\hline
\end{tabular}
\caption{Confusion Matrix summary for Entities in Fold\_1}\label{confusion_matrix_entities}
\end{table}

\end{document}